\documentclass[letterpaper, 10 pt, conference]{ieeeconf}  

\IEEEoverridecommandlockouts                              

\usepackage{graphicx} 

\title{\LARGE \bf
UAV-based Intelligent Information Systems on Winter Road Safety for Autonomous Vehicles

}

\author{Siva Ariram$^{1}$, Veikko Pekkala$^{2}$, Timo Mäenpää$^{1}$, Antti Tikänmaki$^{1}$ and Juha Röning$^{1}$ \\
\authorblockA{\textit{$^{1}$Biomimetics and Intelligent Systems Group,}\\
\textit{Faculty of Information Technology and Electrical Engineering,}\\
\textit{$^{2}$Civil Engineering,}
\textit{Faculty of Technology,}\\
\textit{University of Oulu, Finland}\\
\textit{siva.ariram@oulu.fi}\\}}%

\begin{document}

\maketitle
\thispagestyle{empty}
\pagestyle{empty}


\begin{abstract}


As autonomous vehicles continue to revolutionize transportation, addressing challenges posed by adverse weather conditions, particularly during winter, becomes paramount for ensuring safe and efficient operations. One of the most important aspects of a road safety inspection during adverse weather is when a limited lane width can reduce the capacity of the road and raise the risk of serious accidents involving autonomous vehicles. In this research,  a method for improving driving challenges on roads in winter conditions, with a model that segments and estimates the width of the road from the perspectives of Uncrewed aerial vehicles and autonomous vehicles.  The proposed approach in this article is needed to empower self-driving cars with up-to-date and accurate insights, enhancing their adaptability and decision-making capabilities in winter landscapes.

Keywords - Autonomous Driving, UAV, Digital Elevation Model, Segmentation, Winter Road Inspection, ADAS.


\end{abstract}

\section{INTRODUCTION}

In the realm of autonomous vehicles, ensuring safe and efficient operations in adverse weather conditions, particularly during winter, is a critical challenge. Unpredictable factors such as snow, ice, and low temperatures can significantly impact road conditions, posing challenges for self-driving cars. In response to these challenges, integrating Uncrewed Aerial Vehicles (UAVs) into an intelligent winter road information system offers a promising solution.

This innovative approach leverages UAVs equipped with advanced sensors and imaging technologies to gather real-time data on winter road conditions. The collected information is then transmitted to self-driving cars, empowering them with up-to-date and accurate insights into the environment. Such a system holds the potential to enhance the safety, reliability, and performance of autonomous vehicles navigating through winter landscapes.

This study explores the key components and considerations involved in developing a UAV-based intelligent winter road information system tailored for self-driving cars. From high-resolution imaging and surface weather detection to machine learning algorithms and real-time data transmission, the integration of these elements aims to create a holistic solution that enables autonomous vehicles to navigate winter conditions with heightened awareness and adaptability.

As we delve into the various aspects of this system, we will explore how it addresses challenges related to road safety, decision-making algorithms, regulatory compliance, and emergency response integration. The ultimate goal is to pave the way for a more resilient and responsive autonomous vehicle ecosystem, capable of tackling the complexities posed by winter weather conditions. Through the synergy of UAV technology and self-driving car systems, we aim to unlock new possibilities in the advancement of intelligent transportation, ensuring a safer and more efficient future for autonomous vehicles in winter environments.\\
The following are the primary contributions of this study:
\begin{itemize}
\item The technique and thorough road width assessment in winter conditions rely on the road segmentation and intensity data from the 3D point cloud to determine the discrepancy with the actual road.
\item The method to assess the road during winter conditions relies on the road surface profile to evaluate the quantity of snow heaps on the road.
\item Data Collector for Autonomous Car - Sensor setup provides information about the road conditions, and vehicle's surroundings, including objects, obstacles, and navigation cues.
\end{itemize}

\section{Related Work}

Research and development in the integration of UAVs for intelligent winter road information systems, especially in the context of self-driving cars, is an evolving field with contributions from various domains. Here, we explore some related work and notable studies that have contributed to the understanding and implementation of UAV-based technologies for improving road safety and autonomous vehicle operations in winter conditions.

This comprehensive analysis examines the scientific contributions made to the use of Uncrewed aerial vehicles (UAVs) in civil engineering, particularly in the area of traffic monitoring\cite{butilua2022urban}. It discusses the integration of UAVs with intelligent transportation systems to enhance a positive impact on society by decreasing unpleasant situations like traffic management and collisions in urban cities.


Large-scale data availability has the potential to revolutionize the development of intelligent transportation systems (ITS), transforming them from traditional technology-driven systems into more potent, multi-functional data-driven intelligent transportation systems (D2ITS), which are driven by learning algorithms, vision, and multi-source systems to maximize performance\cite{zhang2011data}. It discusses the integration of high-resolution cameras and thermal imaging to assess road conditions and proposes a system for timely information dissemination to traffic management systems.

To facilitate better autonomous and non-autonomous driving in inclement weather, this research work suggests a technique for assessing road surface slipperiness at the input image pixel level\cite{pesonen2023pixelwise}. By offering a better understanding of road conditions, the high-resolution output information may enable more advancements in fully autonomous driving software, advanced driver assistance systems (ADAS), or human driving through a visual interface or other alert mechanism. Therefore, the system would significantly affect the potential applications of real-time slipperiness data to enhance manual and autonomous driving



This work highlights the potential of UAVs combined with deep learning techniques to offer a more comprehensive view of road conditions, enabling more efficient maintenance solutions including robotic interventions and prefabrication as well as targeted repairs\cite{katsamenis2023real}. Road inspection can be made safer and more affordable by using UAV images in conjunction with the Yolov5 model. This approach lowers the hazards and costs involved with more conventional techniques of inspection.



These works collectively contribute to the understanding of the challenges posed by winter road conditions for autonomous vehicles and highlight the potential of UAV-based technologies in addressing these challenges. As technology continues to advance, the integration of UAVs into intelligent transportation systems for winter road information is expected to play a crucial role in enhancing the safety and efficiency of autonomous vehicle operations.

\section{SYSTEM PLATFORM}

The system platform used in this study consist of a road vehicle equipped with a portable data collector, and an UAV equipped with a camera sensor, GNSS and flight control sensors . The vehicles operated independently of each other during collection of road information data. Collected data was processed offline using a desktop computer. 

\subsection{SMAD Car Sensor}

For collecting data from vehicles perspective a data collector system was mounted in front of the SMAD (Smart and Mobile testbed for Automated and assisted Driving) car. System consists of Intel Realsense D455 stereo camera, Ouster OSDome-128 lidar, Xsens MTi-G-710 GNSS/INS connected to a single board computer. Sensor setup is controlled by in-house developed framework. Received data from the sensor is serialized in Concise Binary Object Representation (CBOR) format and saved to temporary file-system overwriting any previous measurement date from the same sensor. To collect a dataset new files in temporary filesystem (tmpfs) is copied to solid state drive on the controlling computer. When copying files are renamed to add timestamp and sensor id to filename. Dataset is a collection of files that can be sorted by timestamp and easily divide to smaller time series.

\begin{figure}[thpb]
\centering

\includegraphics[width=3.4in]{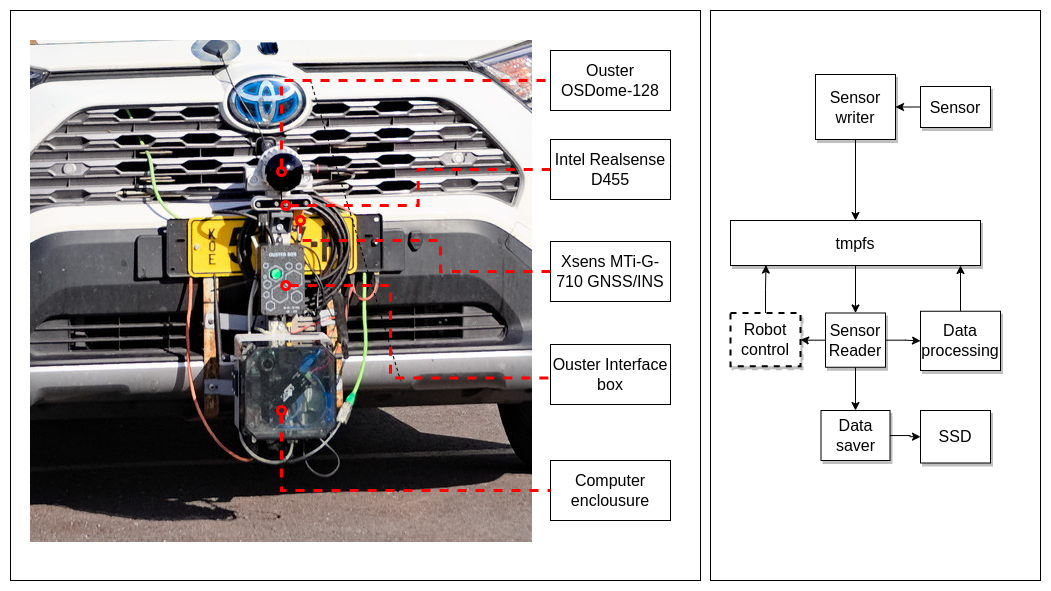}

\caption{(LHS) Sensor setup mounted in front of the vehicle with sensor labels.  (RHS) Data flow in the framework. Raw data is read from the sensor, converted to SI-units and saved to tmpfs by sensor writer. Data is then read from temporary file system using sensor reader object and processed for future use. Processed data is saved back in the tmpfs and can be read by next part of the processing pipeline. Data in the tmpfs can be utilized for real time robot control or saved for offline use by data saver.}
\label{fig:frameworkdataflow }
\end{figure}

\subsection{UAV sensor system}
\begin{figure}[thpb]
\centering
\includegraphics[width=3.4 in]{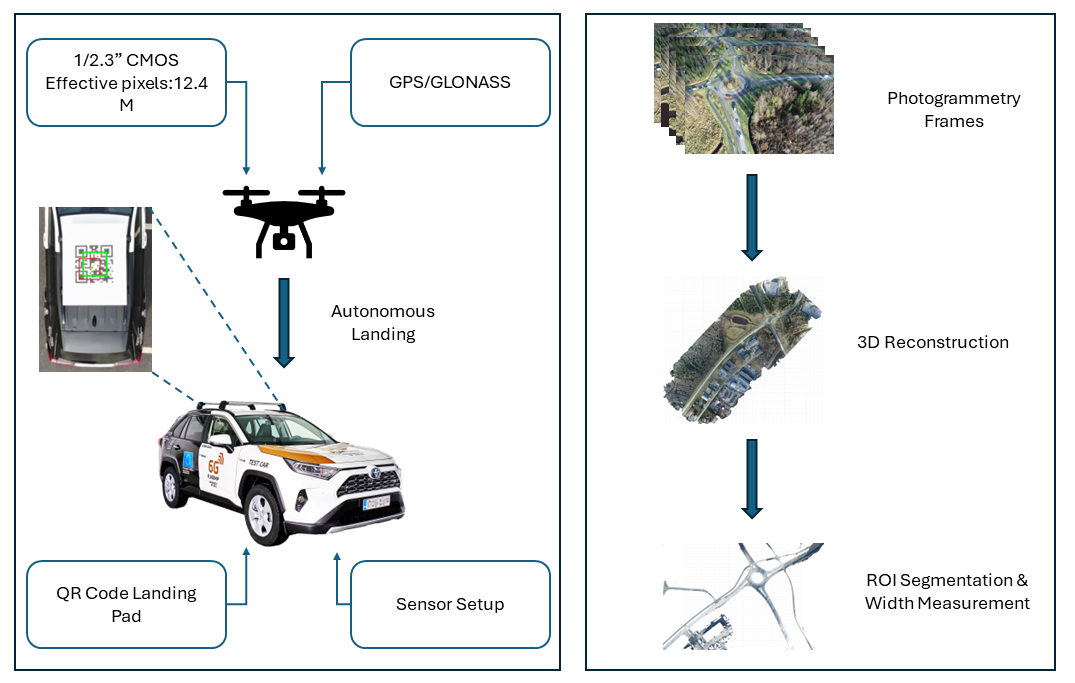}
\caption{UAV Sensor system architecture}
\label{UAV_sensor}
\end{figure}

The information from the UAV sensor system perspective provides a more robust and safe autonomous driving system to the users during navigation. Measurements of the actual road dimensions in adverse weather can also be made using the mobility data that the UAV sensor unit provides. Furthermore, this data can be contrasted from the autonomous vehicle's point of view. The purpose of the camera sensors installed in the UAV is to deliver photogrammetry data in three dimensions regarding the area of interest. Better information regarding the heading direction of the
autonomous vehicle can be obtained from this UAV data. Figure \ref{UAV_sensor} shows the data processing and collection process using various sensor modules in autonomous vehicles and UAVs. Wherein Agisoft Metashape Professional is used to create the 3D model, digital elevation model, and orthomosaic images. In addition, it is published in Pointscene so that the difference between the actual road and adverse weather can be observed.

\section{Road Inspection Properties}

Road detection is a crucial component of computer vision systems used in various applications, including autonomous vehicles, advanced driver-assistance systems (ADAS), and smart city infrastructure. Detecting the road accurately is essential for navigation, lane-keeping, and ensuring overall safety. Several methods are employed for road detection, and these can be categorized into traditional computer vision techniques and more advanced deep learning approaches. The lane detection through the combination of edge detection, Hough transform, and curvature analysis to identify lane boundaries\cite{javeed2023lane}. It recognizes only lines that are straight. Determining the curvature of lanes, or curved road lanes, is a challenging problem.

\subsection{Road Detection Methods}

The choice of a road detection method depends on factors such as the complexity of the environment, available sensor modalities, computational resources, and real-time requirements. In many contemporary applications, deep learning-based methods, particularly semantic segmentation networks\cite{pennanen2021all}, have shown significant success due to their ability to automatically learn complex features and patterns from large datasets.

\subsection{Road Structure and Road Cross-sections}


The road structure refers to the physical and functional elements that make up a road, encompassing various components designed to facilitate safe and efficient transportation. The structure of a road includes both the visible elements, such as travel lanes, signs, and markings, and the underlying infrastructure that supports road functionality. The key components are roadway, markings and signs, roadside elements, and Traffic Management systems (Intelligent Transportation Systems (ITS)). Understanding the road structure is essential for road design, traffic management, and ensuring the safety of all road users. For the purposes of this study, the essential aspect to understand about road structure is the road cross-section and its various elements and their dimensions.

\begin{figure}[thpb]
\centering
\includegraphics[width=3.4 in, height= 2 in]{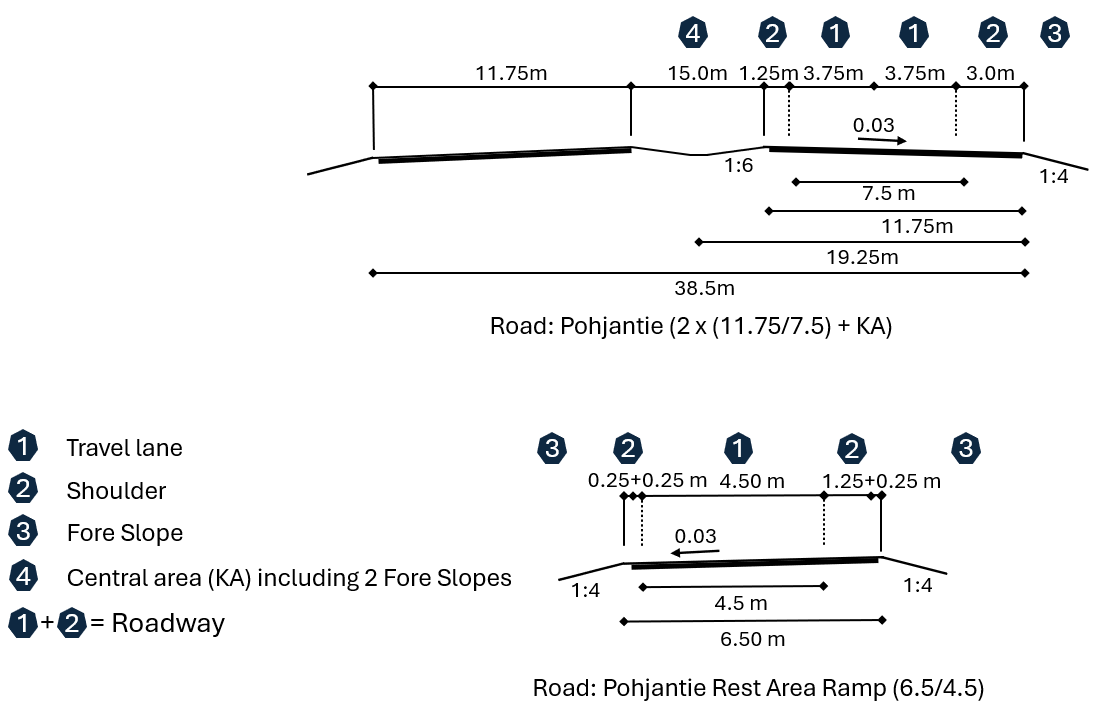}
\caption{Road Cross Section Design Elements in Pohjantie}
\label{Pohjantie_CS}
\end{figure}

The road cross-section is made up of sub-elements, the use and dimensions of which are determined by the road classification, design traffic volume, design traffic composition, design speed and the desired quality class. For example, the performance of main arterial roads is particularly affected by weekend and summer peak hours and heavy weekday traffic, while on the lower road network, the choice of road cross-section is influenced by the volume of both heavy and pedestrian and cyclist traffic. Desired quality class refers to usage of indicative cross-sections or minimum level cross-sections\cite{repo2023vaylahankkeen}.
\begin{figure}[thpb]
\centering
\includegraphics[width=3.4 in, height= 2 in]{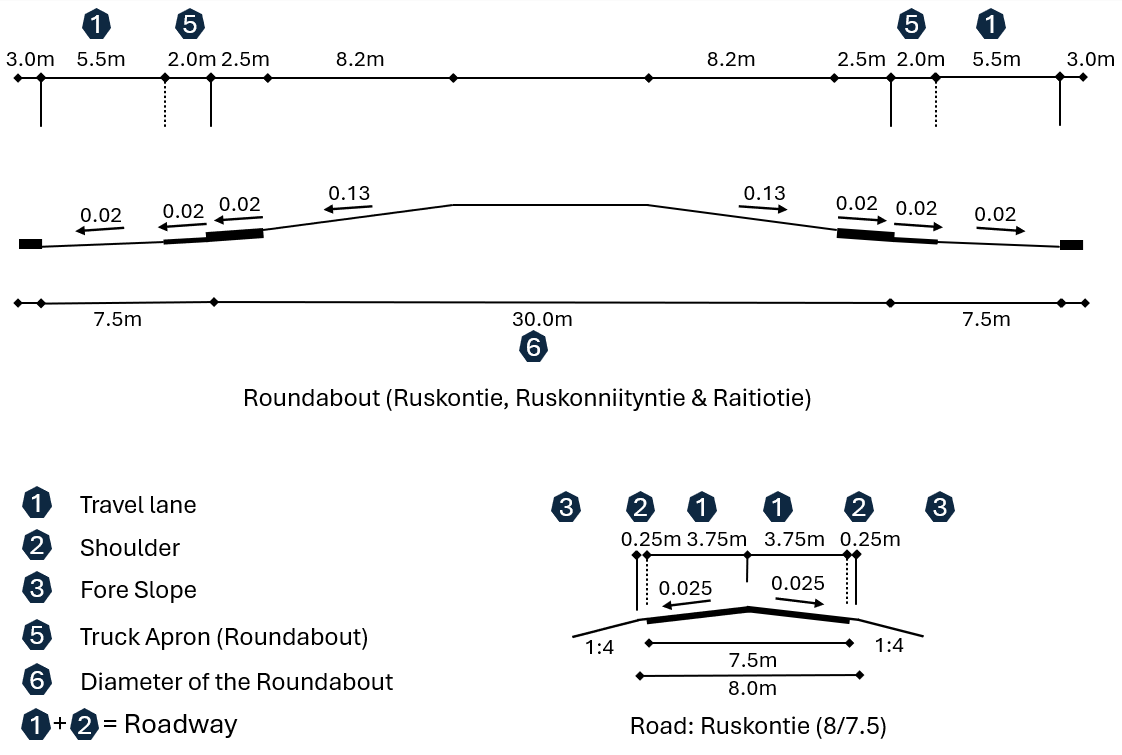}
\caption{Roundabout and Road Cross Section Design Elements in Ruskontie}
\label{Rusko_CS}
\end{figure}

Road cross-sections are described by symbols indicating the number of travel lanes on two-lane road and, on all cross-sections, the combined width of the travel lanes and shoulders and the width of the travel lanes separated by a slash (/). The label also indicates the type of separation of the two-lane road: central area (KA), steel guardrail 
(TK) or concrete railing (BK)\cite{repo2023vaylahankkeen}.

Figures \ref{Pohjantie_CS} and \ref{Rusko_CS} presents roads under investigation. In Figure \ref{Pohjantie_CS} road cross-section design elements area presented for Pohjantie Highway and Pohjantie Rest Area Ramp. Design speed for the Pohjantie highway is 100 km/h, in year 2021 the average amount of vehicles per day was 15 860 including 1101 heavy trucks. Pohjantie highway is designed as two roadways (both having two travel lanes for both directions) separated with central area. Cross-section design elements symbols are 2 x (11.75/7.5) + KA. In Figure \ref{Rusko_CS} road cross-section design elements area presented for Ruskontie Urban Road and Roundabout for 3 roads (Ruskontie, Ruskonniityntie and Raitiotie). Ruskontie is main regional road. Design speed for the road is 60 km/h, in year 2021 average amount of vehicles per day was 4 300 including 215-430 heavy trucks. Ruskontie is designed as roadway with two travel lanes (one for both directions). Cross-section design elements symbols are (8/7.5). For the roundabout average amount of vehicles per day in 2021 was 14 200. 

\begin{figure*}[ht!]
\begin{center}
\includegraphics[width=7in, height = 2.5in]{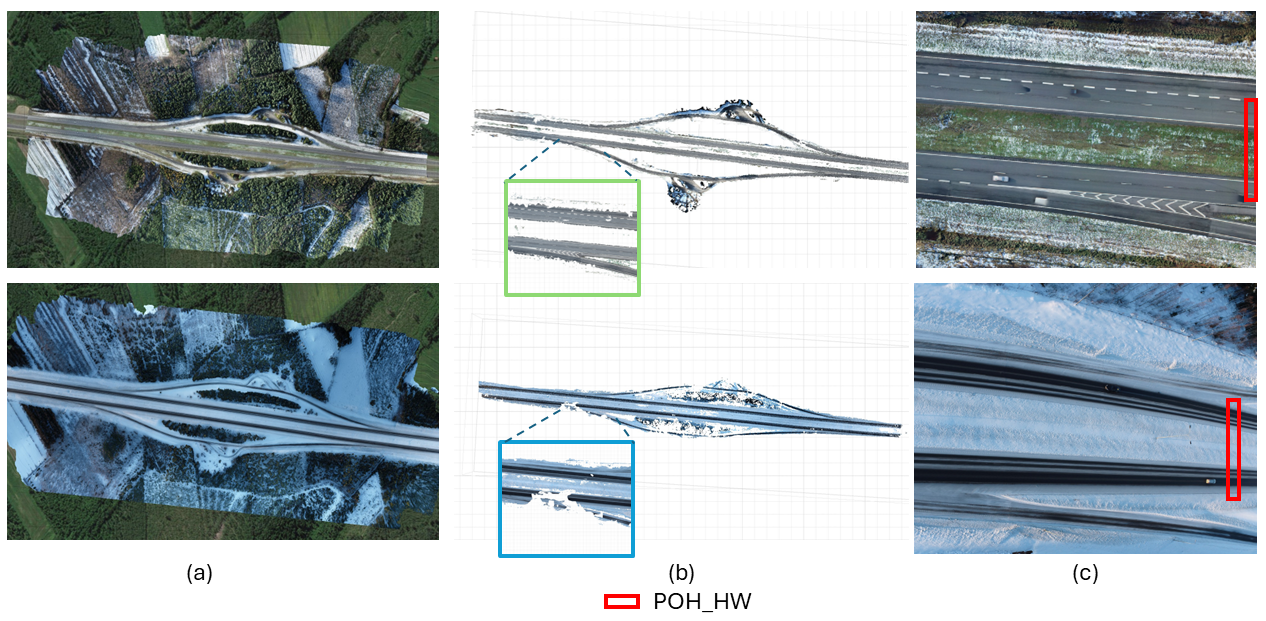}
\caption{Pohjantie Highway 38.5m: (a) 3D Generated colored dense cloud, (b) 3D ROI Segmented Road surface, (c) 2D ROI UAV Ortho images}
\label{fig_poh_reconstruction360}
\end{center}
\end{figure*}

From the investigation point of view most important road cross-sections and dimensions are Travel line (1), Shoulder (2), Fore Slope (3), and Roadway (1+2). For roundabout also Truck Apron (4) and  Diameter for Roundabout (5) should be mentioned. Travel line (1) is portion of the road marked or otherwise designated for motor vehicles, with sufficient width in the longitudinal direction of the roadway, or bicycle lane. Shoulder (2) is longitudinal part of the road separated from the travel lane by a boundary line. The shoulder is divided into paved shoulder and gravel shoulder. The purpose of the shoulder is to support the driving lane, enhance traffic safety and road capacity, and serve as a temporary parking space for broken-down vehicles. The shoulder can also serve as a part of the road for pedestrian and bicycle traffic as needed. Roadway (1+2) is road section designated for motor vehicle traffic, consisting of one or more travel lanes, excluding bicycle lanes.\cite{repo2023vaylahankkeen}\cite{tiehallinto2021}

\subsection{Data Evaluation and DEM Generation}
Data evaluation involves assessing the quality, reliability, and relevance of data collected for a particular purpose. In the context of UAV-based intelligent information systems for winter road safety in autonomous vehicles, data evaluation plays a crucial role in ensuring the effectiveness and accuracy of the system. The Table \ref{data_table} shows the dataset that we gathered included data from the winter and summer seasons in the targeted areas, which included Pohjantie Highway Road and Ruskontie Urban Road. The surface areas covered by the data set are 99076.9 and 14711.3m\textsuperscript{2}, respectively. Driving in the AV, the average speed ranges from 50 to 120 Kmph, while the average flying speed when gathering data from the UAV is 7.1 M/s. With the timed interval shoot mode enabled and the gimbal angle set to -60\textsuperscript{o}, the height buffer for surface extraction from the UAV was 90 meters.

In order to obtain the maximum folded points possible in a 3D model, we used photogrammetry flying method, which can overlap 80\% in both the vertical and horizontal directions\cite{haala2012performance}. The technique of georeferencing involves assigning precise geographic coordinates, usually given in latitude and longitude, to spatial data, such as images or maps. To utilize the data in a geomatic application, georeferencing (ETRS89/ETRS-GK26FIN) is required\cite{uysal2015generation}. Inertial time of the camera and GPS time must coincide for direct georeferencing to work. After making these alterations to the data, the 3D reference point cloud and digital elevation model can be generated directly. 
\begin{table}[h]
\caption{Segments of roads gathered for this study}
\label{data_table}
\begin{center}
\begin{tabular}{|p{0.42in}||p{0.43in}||p{0.95in}||p{0.4in}||p{0.22in}|}
\hline
\textbf{Segment of Road} & \textbf{Collection Date} & \textbf{Sensors Used}& \textbf{Type of Road} & \textbf{No.of Lanes}\\
\hline
Pohjantie & 10.10.2023\newline(Autumn) \newline 01.09.2024\newline(Winter) & Intel Realsense D455 \newline OSDome-128 \newline MTi-G-710 GNSS/INS\newline  1/2.3” CMOS \newline GPS/GLONASS & Sate Roads& 4\\
\hline
\hline
Ruskontie & 10.10.2023\newline(Autumn) \newline 01.09.2024\newline(Winter) & Intel Realsense D455 \newline OSDome-128 \newline MTi-G-710 GNSS/INS\newline  1/2.3” CMOS \newline GPS/GLONASS& Municipal roads & 2\\
\hline
\end{tabular}
\end{center}
\end{table}

\section{Experimentation Results}
The progress of autonomous vehicle winter road safety solutions is aided by the valuable insights that trial outcomes in this domain offer on the functionality and efficacy of UAV-based intelligent information systems. The experimentation results includes assessing the ability of UAV-based systems to detect hazards on winter roads, such as snow heaps and lane width. We believe that the assessment impacts the UAV-based intelligent information systems on the performance of autonomous vehicles operating in winter conditions.   

\subsection{Road Width Estimation and Segmentation from UAV images}
\begin{figure}[thpb]
\centering
\includegraphics[width=3.4 in, height= 1.75 in]{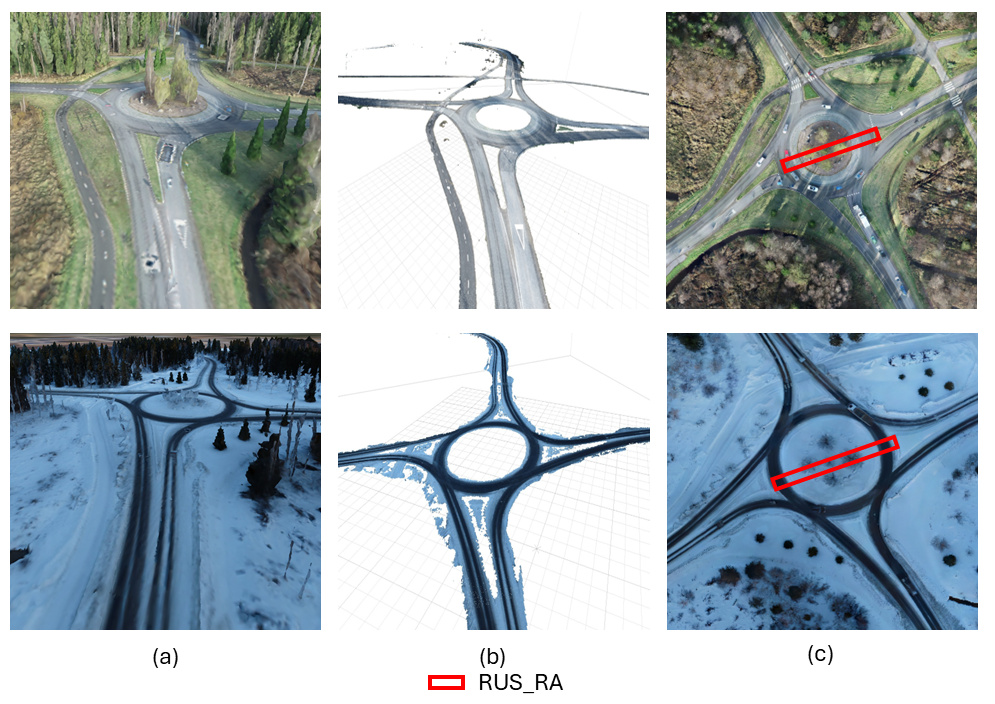}
\caption{Ruskontie Urban Roadway - Roundabout: (a) 3D Generated colored dense cloud, (b) 3D ROI Segmented Road surface, (c) 2D ROI UAV Ortho images}
\label{Rusko_RA}
\end{figure}

\begin{figure}[thpb]
\centering
\includegraphics[width=3.4 in, height= 1.75 in]{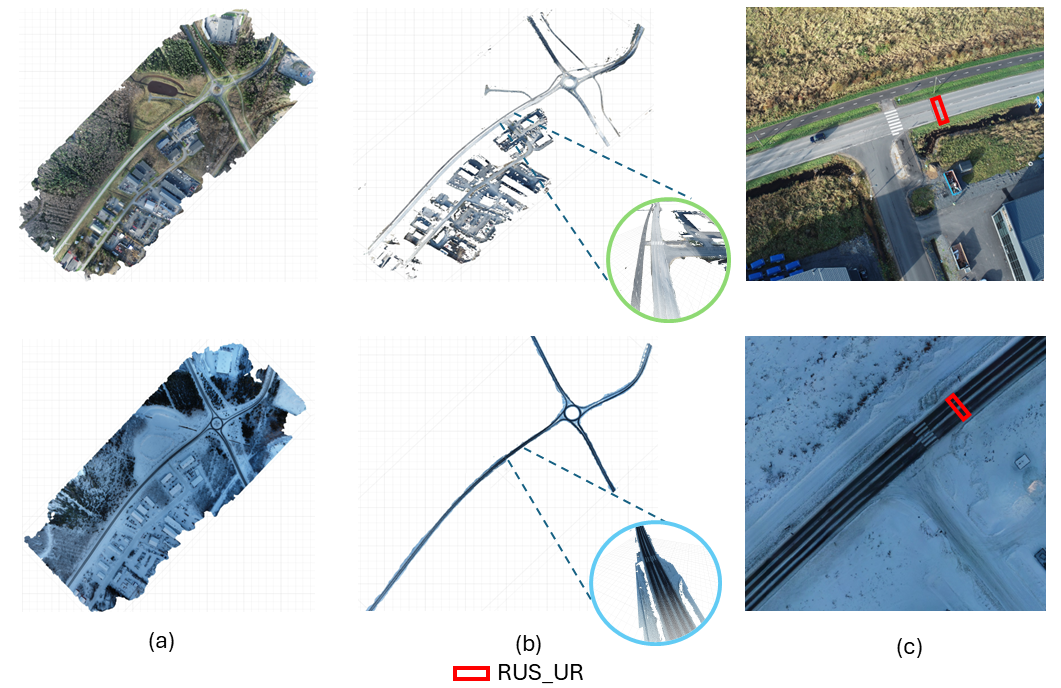}
\caption{Ruskontie Urban Roadway 7.5m: (a) 3D Generated colored dense cloud, (b) 3D ROI Segmented Road surface, (c) 2D ROI UAV Ortho images}
\label{Rusko_UR}
\end{figure}

Estimating road width is a fundamental task in autonomous driving. Road width estimation typically involves analyzing images or sensor data to determine the width of the road surface accurately. Using the data gathered in the study area, the road segmentation analysis was carried out on dense clouds that were generated. The corresponding region of interest for each segmentation is depicted in Figures \ref{fig_poh_reconstruction360}, \ref{Rusko_RA}, and \ref{Rusko_UR}. Because of the snow heaps on the road surface, it is evident in Figure \ref{fig_poh_reconstruction360}(b) that the segmentation algorithm\cite{pennanen2021all} was unable to identify the asphalt. Similarly, the approach failed to identify the complete road division which is turning right in Figure \ref{Rusko_UR}(b). This indicates the deficiency of segmentation models, especially for segmenting roads during the winter.  

\begin{figure}[thpb]
\centering
\includegraphics[width=3.4 in]{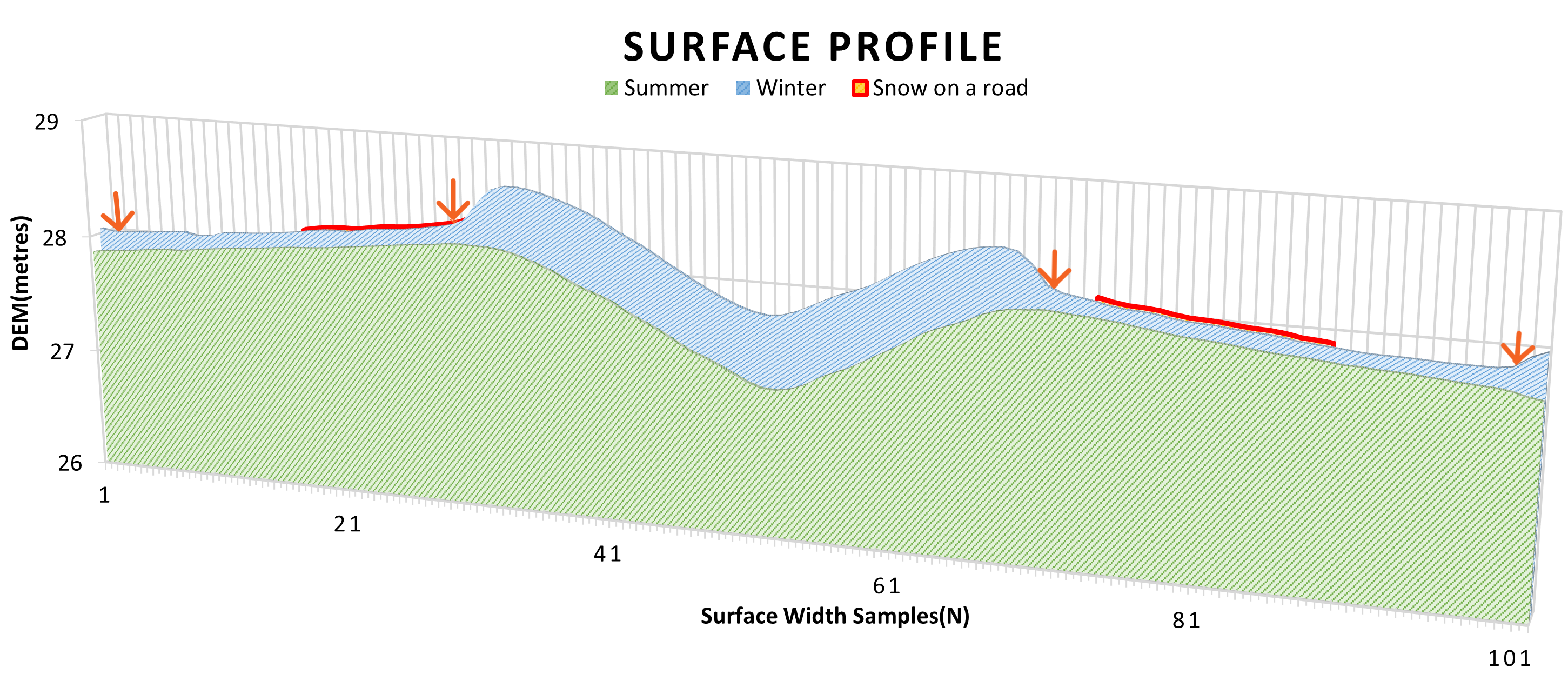}
\caption{Pohjantie Highway Surface Profile, Driveway markers are indicated with an orange arrow.}
\label{Pohjantie_Graph}
\end{figure}

\begin{figure}[thpb]
\centering
\includegraphics[width=3.4 in]{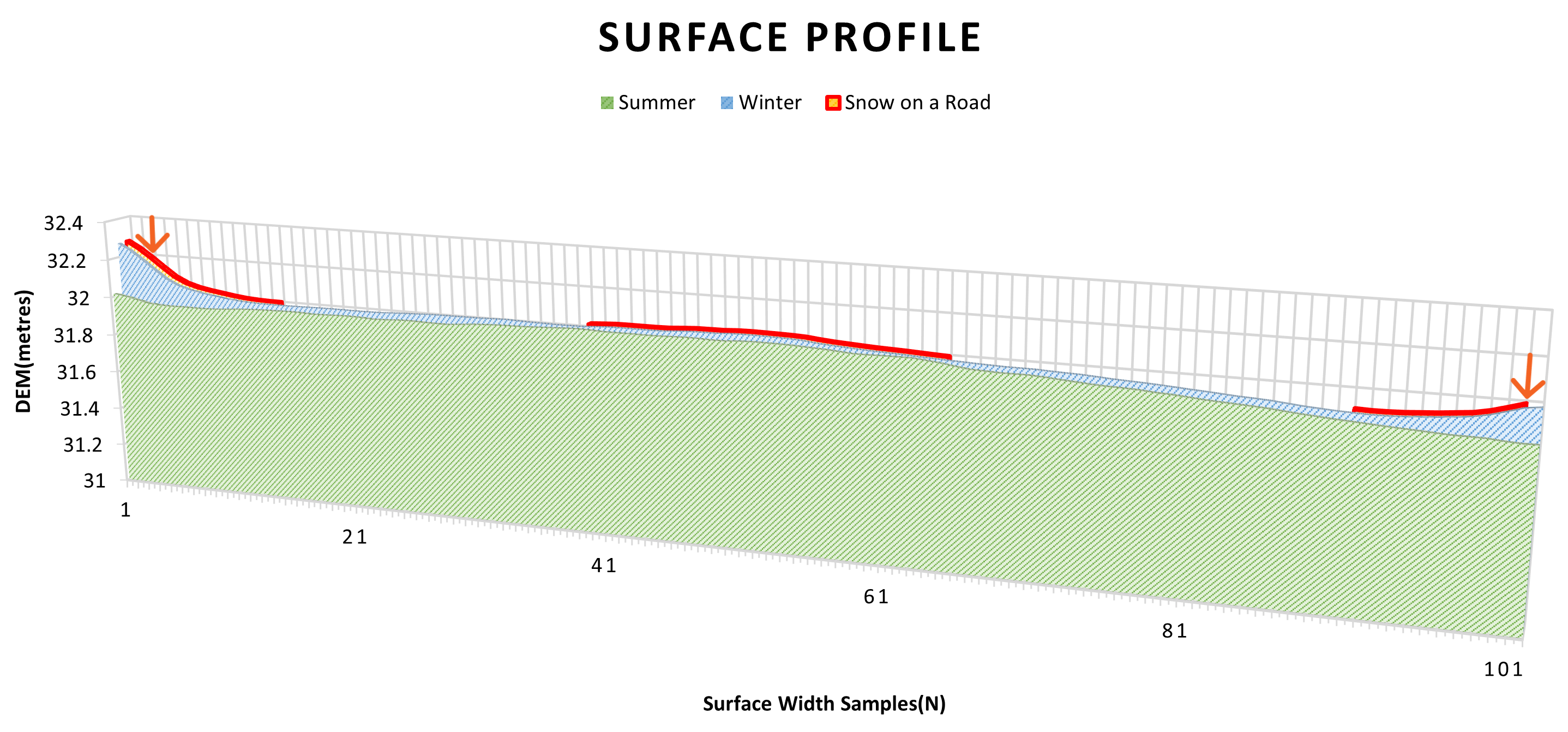}
\caption{Ruskontie Urban Roadway Surface Profile, Driveway markers are indicated with an orange arrow.}
\label{Ruskontie_road_Graph}

\end{figure}
\begin{figure}[thpb]
\centering
\includegraphics[width=3.4 in]{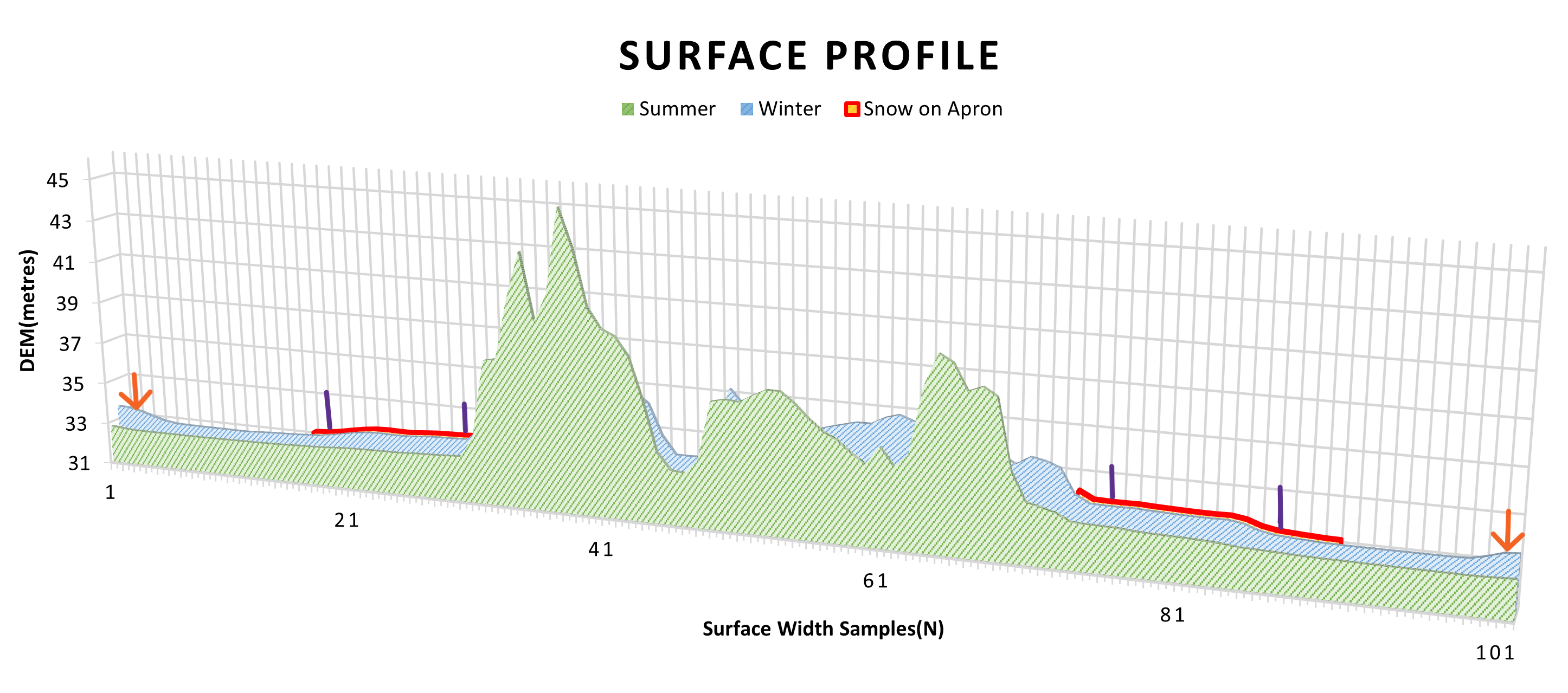}
\caption{Ruskontie Roundabout Surface Profile, Driveway markers are indicated with an orange arrow and Aprons are indicated with a violet line.}
\label{Ruskontie_Rb_Graph}
\end{figure}

\subsection{Road Surface Measurement - UAV Perspective}
A precise and unambiguous characterization of the nominal surface—that is, the actual surface that is meant to be modeled—is necessary to evaluate the road surface from the DEM\cite{polidori2020digital}. When comparing a digital elevation model (DEM) obtained through photogrammetry or lidar, where the measured elevation lies at the top of the snow, with ground-based control points, one may mistakenly conclude that the DEM and the mapping method are not accurate because the reference data do not represent the same surface as the DEM. Therefore, without a precise definition of the actual surface that a DEM is meant to reflect, it would be pointless to assess the quality of a DEM.

Based on the assessment, we were able to determine that the surface profile in Figure \ref{Pohjantie_Graph} even displays the smaller snow heaps on the inside lanes, which are referred to as overtaking lanes and are hardly used (Indicated by red markings on the horizontal axis). In Figure \ref{Pohjantie_Graph} the Pohjantie Highway is divided into two roads with a central area in between, each with two lanes of traffic in either direction. We discovered that there was a significant variation in the width of the road, which had an impact on the cross-section design features (The orange arrow marking shows the end of the driveway). In Figure \ref{Ruskontie_road_Graph} Ruskontie Urban roadway is designed for two lanes in either direction, the graph displays a heap of snow in the middle of the road, as well as snow covering the entire shoulder and a portion of the driving lane (Indicated by red markings on the horizontal axis). In Figure \ref{Ruskontie_Rb_Graph} the Ruskontie roundabout features four exit options and is only intended for one lane in one direction. The graph displays the amount of snow that has been collected on the truck apron and the portion of the road. Snow has entirely covered the inner shoulder (Indicated by red markings on the x-axis).

\subsection{Road Surface Measurement - Autonomous Car Perspective}  



\begin{figure}[thpb]
\centering
\includegraphics[width=3.4 in ]{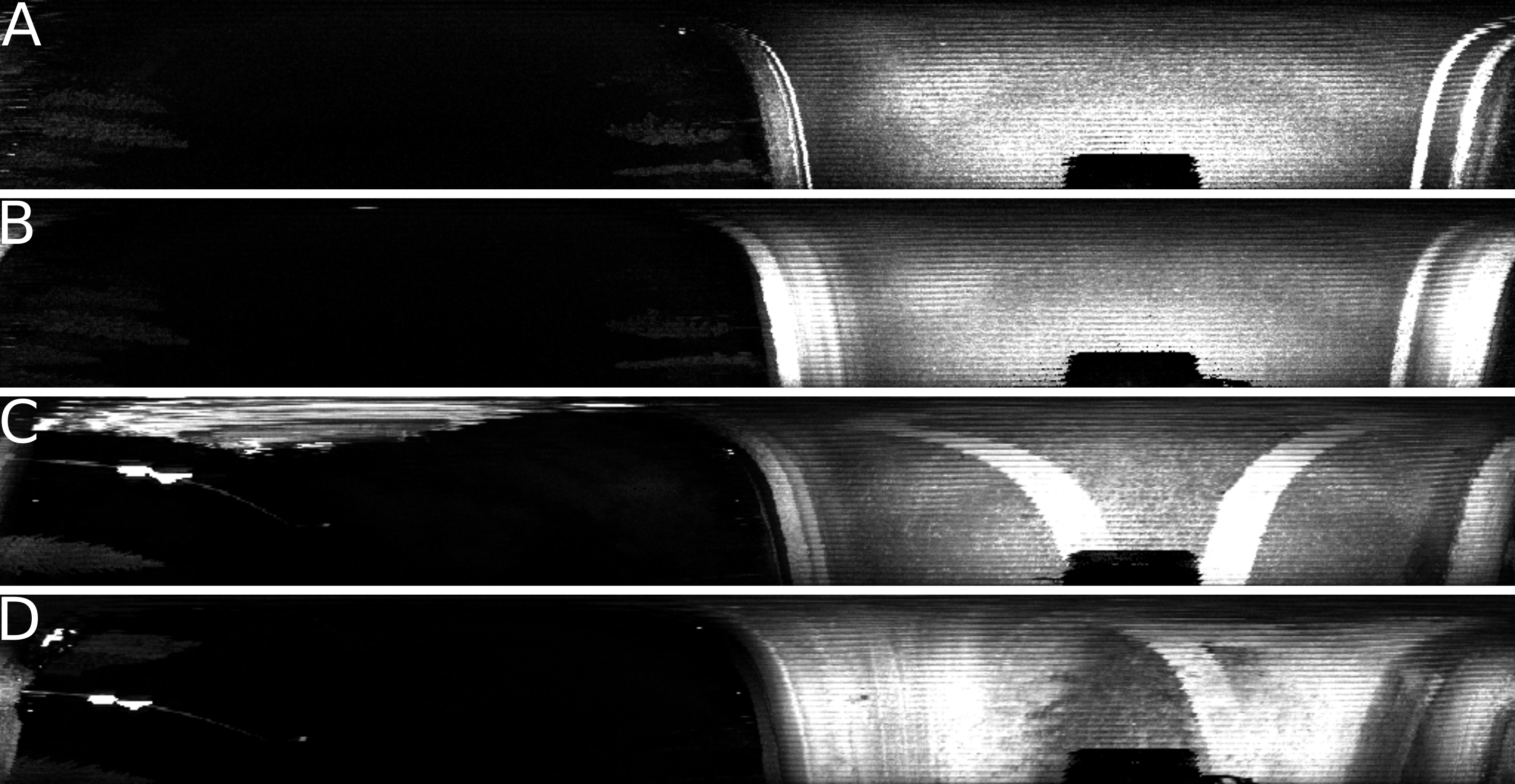}
\caption{Intensity images form Ouster OS-Dome lidar during October and January with road markings and snow showing higher intensities. panoramic Intensity images were generated from recorded data by normalizing and destaggering raw intensity measurements using a simple python script. (A) Highway exit on Pohjantie Highway showing lane marking on October (B) Same exit on January with snow on the edges of road covering left side lane markings (C) Ruskontie Urban Roadway - Roundabout during October showing yielding triangle painted on road. (D) Same roundabout during January with snow partially obscuring road markings. }
\label{lidar_intesities}
\end{figure}

Portable sensor system mounted in front of the car is near the road surface and has clear view of the road surface enabling detection of road features like road markings and unevenness. Reflective road markings or patches of ice and snow can be detected using lidar using intensity values from the measurement. Lidar images  visualizing intensity values of road features are presented in Figure \ref{lidar_intesities}. During January road markings are partially or completely covered by snow making it more difficult to detect lanes. 

\begin{figure}[thpb]
\centering
\includegraphics[width=3.4 in ]{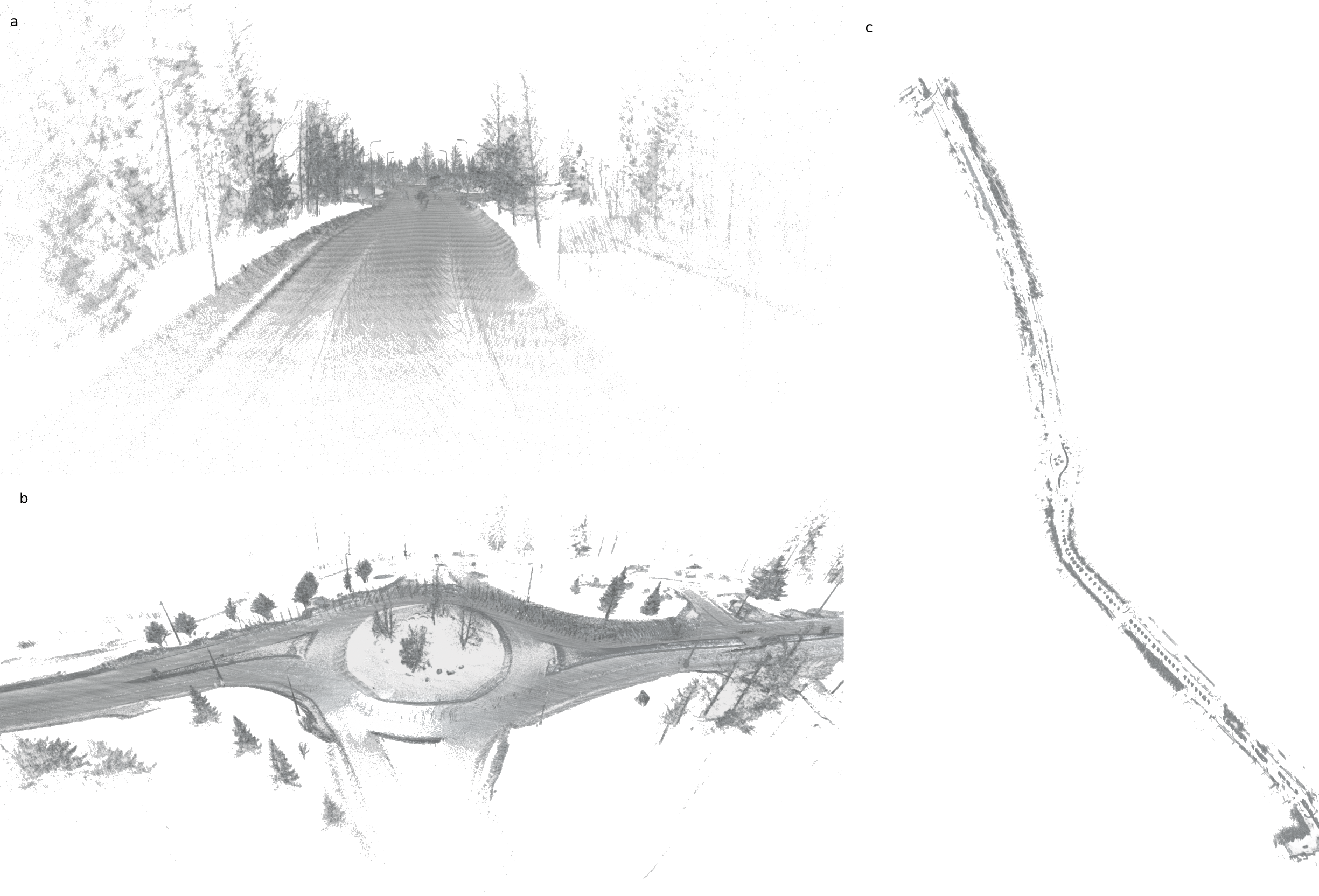}
\caption{Generated lidar pointclouds from Ruskontie Urban Roadway (a) Road surface with visible snow-banks on both sides (b) Roundabout (c) Overview of the map }
\label{lidar_pointcloud}
\end{figure}

Using data collected during January a 3D-map of Ruskontie was constructed using KISS-ICP\cite{vizzo2023kiss} and SHINE mapping\cite{zhong2023shine}. In Figure \ref{lidar_pointcloud} images from generated point cloud are presented showing visible snow-banks in the edges of the road. From the generated data it is possible to estimate the width of the road using visible snow-banks at the edges of the road even if lane markings are covered by snow. Using estimated width and knowledge of number of lanes lane detection could be possible.

\section{CONCLUSIONS}

In conclusion, the integration of Uncrewed Aerial Vehicles (UAVs) into intelligent information systems represents a promising solution for enhancing winter road safety in autonomous vehicles (AVs). Throughout this paper, we have highlighted the importance of addressing the challenges posed by adverse weather conditions, particularly during winter, to ensure the reliable operation of AVs. By leveraging UAVs equipped with advanced sensors and imaging technologies, real-time data on winter road conditions can be collected and transmitted to AVs. This enables AVs to make informed decisions, adapt their driving behavior, and mitigate risks associated with snow, ice, and reduced visibility. With continued innovation and investment in this area, we can pave the way for a future with safer, more resilient, and weather-adaptive autonomous transportation systems.

\addtolength{\textheight}{-12cm}   




\section*{ACKNOWLEDGMENT}

We are grateful to our colleagues and collaborators who have provided support and feedback throughout the research process, helping to refine our ideas and improve the quality of the paper. Additionally, we acknowledge the support of our institutions and funding from the European Union's Horizon 2020 \textbf{Flying Forward 2020} (Grant Agreement no.101006828) and \textbf{SMAD} (Project Code. A74419) that have provided resources and support for this research endeavor.


\begin{thebibliography}{1}

\bibitem{butilua2022urban} Butilă, E. V., \& Boboc, R. G. (2022). Urban traffic monitoring and analysis using unmanned aerial vehicles (uavs): A systematic literature review. Remote Sensing, 14(3), 620.

\bibitem{zhang2011data} Zhang, J., Wang, F. Y., Wang, K., Lin, W. H., Xu, X., \& Chen, C. (2011). Data-driven intelligent transportation systems: A survey. IEEE Transactions on Intelligent Transportation Systems, 12(4), 1624-1639.
\bibitem{pesonen2023pixelwise} Pesonen, J. (2023). Pixelwise Road Surface Slipperiness Estimation for Autonomous Driving with Weakly Supervised Learning.
\bibitem{katsamenis2023real}Katsamenis, I., Bakalos, N., Protopapadakis, E., Karolou, E. E., Kopsiaftis, G., \& Voulodimos, A. (2023, July). Real time road defect monitoring from UAV visual data sources. In Proceedings of the 16th International Conference on PErvasive Technologies Related to Assistive Environments (pp. 603-609).
\bibitem{javeed2023lane}Javeed, M. A., Ghaffar, M. A., Ashraf, M. A., Zubair, N., Metwally, A. S. M., Tag-Eldin, E. M., ... \& Jiang, X. (2023). Lane Line Detection and Object Scene Segmentation Using Otsu Thresholding and the Fast Hough Transform for Intelligent Vehicles in Complex Road Conditions. Electronics, 12(5), 1079.
\bibitem{pennanen2021all} Pennanen, T., Ariram, S., Tikanmäki, A., \& Röning, J. (2021, August). All-Around 3D Reconstruction from Spherical Images with Semantic Segmentation. In 2021 IEEE International Conference on Mechatronics and Automation (ICMA) (pp. 193-199). IEEE.
\bibitem{haala2012performance}Haala, N., Cramer, M., Weimer, F., \& Trittler, M. (2012). Performance test on UAV-based photogrammetric data collection. The International Archives of the Photogrammetry, Remote Sensing and Spatial Information Sciences, 38, 7-12.
\bibitem{uysal2015generation} Uysal, M., Toprak, A. S., \& Polat, N. (2015). DEM generation with UAV Photogrammetry and accuracy analysis in Sahitler hill. Measurement, 73, 539-543.
\bibitem{repo2023vaylahankkeen} Repo, E. (2023). Väylähankkeen suunnittelun nykytilakartoitus.
\bibitem{tiehallinto2021}Tasoliittym{\"a}t. Suunnitteluvaiheen
ohjaus (2001)
\bibitem{vizzo2023kiss}Vizzo, I., Guadagnino, T., Mersch, B., Wiesmann, L., Behley, J., \& Stachniss, C. (2023). Kiss-icp: In defense of point-to-point icp–simple, accurate, and robust registration if done the right way. IEEE Robotics and Automation Letters, 8(2), 1029-1036.
\bibitem{zhong2023shine}Zhong, X., Pan, Y., Behley, J., \& Stachniss, C. (2023, May). Shine-mapping: Large-scale 3d mapping using sparse hierarchical implicit neural representations. In 2023 IEEE International Conference on Robotics and Automation (ICRA) (pp. 8371-8377). IEEE.
\bibitem{polidori2020digital}Polidori, L., \& El Hage, M. (2020). Digital elevation model quality assessment methods: A critical review. Remote sensing, 12(21), 3522.



\end{thebibliography}
\end{document}